\documentclass[10pt,twocolumn,letterpaper]{article}

\usepackage{cvpr}
\usepackage{times}  
\usepackage{helvet}  
\usepackage{courier}  
\usepackage{url}  
\usepackage{graphicx}  

\usepackage{epsfig}
\usepackage{graphicx}
\usepackage{amsmath}
\usepackage{amssymb}
\usepackage{color}
\usepackage{multirow}
\usepackage{caption}
\usepackage{subcaption}
\usepackage{tabularx}
\usepackage{algorithm}
\usepackage{algorithmic}
\usepackage{flushend}
\usepackage{booktabs}

\usepackage[pagebackref=true,breaklinks=true,letterpaper=true,colorlinks,bookmarks=false]{hyperref}

\cvprfinalcopy 

\def\cvprPaperID{216} 

\ifcvprfinal\fi
\begin{document}

\title{Learning Generalizable and Identity-Discriminative Representations for \\ Face Anti-Spoofing}

\author{\normalsize{Xiaoguang~Tu$^{1}$, Jian~Zhao$^{2,3}$, Mei~Xie$^{1}$, Guodong~Du$^{2}$, Hengsheng~Zhang$^{1}$, Jianshu~Li$^{2}$, Zheng~Ma$^{1}$, Jiashi~Feng$^{2}$} \\
	\small{$^{1}$University of Electronic Science and Technology of China, $^{2}$National University of Singapore} \\
		\small{$^{3}$National University of Defense Technology} \\
	{\small  xguangtu@outlook.com, zhaojian90@u.nus.edu, mxie@uestc.edu.cn, \{duguodong7, zhsaimi\}@gmail.com} \\ {\small jianshu@u.nus.edu, elefjia@nus.edu.sg}}
\maketitle

\begin{abstract}
Face anti-spoofing (\textit{a.k.a} presentation attack detection) has drawn growing attention due to the high security demand in face authentication systems. Existing CNN-based approaches usually well recognize the spoofing faces when training and testing spoofing samples display similar patterns, but their performance would drop drastically on testing spoofing faces of unseen scenes. In this paper, we try to boost the generalizability and applicability of these methods by designing a CNN model with two major novelties. First, we propose a simple yet effective Total Pairwise Confusion (TPC) loss for CNN training, which enhances the generalizability of the learned Presentation Attack (PA) representations. Secondly, we incorporate a Fast Domain Adaptation (FDA) component into the CNN model to alleviate negative effects brought by domain changes. Besides, our proposed model, which is named Generalizable Face Authentication CNN (GFA-CNN), works in a multi-task manner, performing face anti-spoofing and face recognition simultaneously. Experimental results show that GFA-CNN outperforms previous face anti-spoofing approaches and also well preserves the identity information of input face images.
\end{abstract}


\section{Introduction}
Despite the recent noticeable advances, the security of face recognition systems is still vulnerable to Presentation Attacks (PA) with printed photos or replayed videos. To counteract PA, face anti-spoofing  \cite{ij:tan2010face,ij:liu2018learning} is developed and serves as a pre-step prior to face recognition.
\begin{figure}[t!]
    \hspace{0.0cm}
    \includegraphics[width=8.5cm, height=3.1cm]{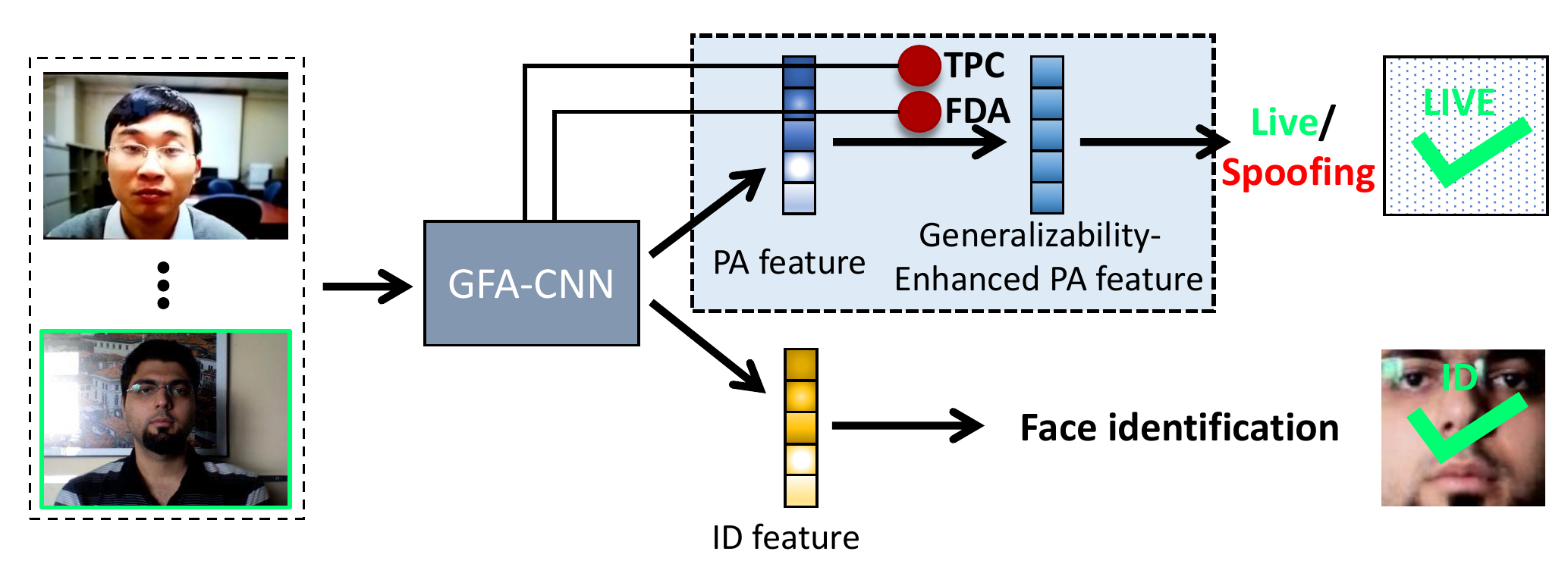}%
   \vspace{-0.3cm}
   \caption{Our CNN framework works in a multi-task manner, addressing face recognition and face anti-spoofing at one shot. It leverages total pairwise confusion (TPC) loss and fast domain adaption (FDA) to enhance the generalizability of the learned Presentation Attack (PA) feature and improve face anti-spoofing performance across different scenes.}
   \vspace{-0.3cm}
\label{fig:0}
\end{figure}

\begin{figure*}[t!]
    \hspace{0.33cm}
    \includegraphics[width=17cm, height=4.7cm]{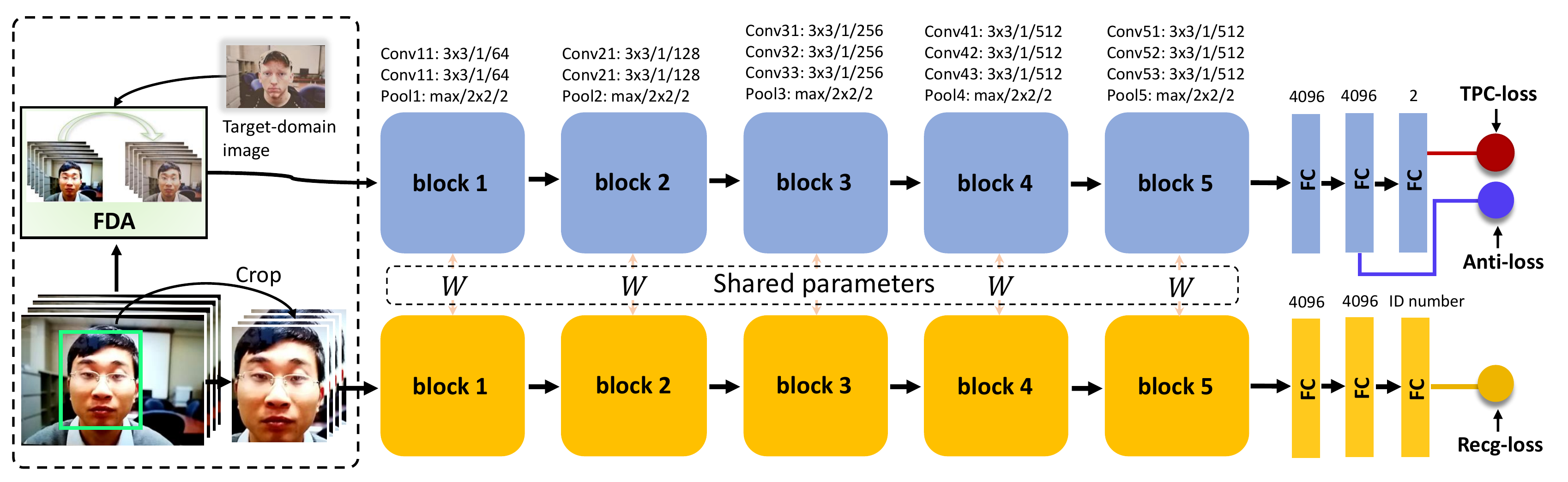}%
   \vspace{-0.2cm}
   \caption{Architecture of proposed GFA-CNN.  The whole network contains two branches. The face anti-spoofing branch (upper) takes as input the domain-adaptive images transferred by FDA and optimized by TPC-loss and Anti-loss, while the face recognition branch (bottom) takes the cropped face images as input and is trained by minimizing Recog-loss.  The structure settings are shown on top of each block, where ``ID number'' indicates the number of subjects involved in training. The two branches share parameters during training.}
   \vspace{-0.3cm}
\label{fig:1}
\end{figure*}

Earlier face anti-spoofing approaches mainly adopt handcrafted features, like LBP \cite{ij:chingovska2012effectiveness}, HoG \cite{ij:komulainen2013context} and SURF \cite{ij:boulkenafet2017face}, to find the differences between live and spoofing faces. In \cite{ij:yang2014learn}, CNN was used for face anti-spoofing for the first time, with remarkable performance achieved in intra-database tests. Following their work, a number of CNN-based methods have been proposed, almost all  treating face anti-spoofing as a binary (live \textit{vs}. spoofing) classification problem. However, given the enormous solution space of CNN, these methods tend to suffer overfitting and poor generalizability to new PA patterns and environments. In this work, we attempt to enable an anti-spoofing system to be deployed in various environments, i.e. with good generalizability.

For CNN-based methods, an important clue to differentiate live \textit{vs.} spoofing faces is the \textit{spoof pattern}, including color distortion, moir$\acute{\text{e}}$ pattern, shape deformation, spoofing artifacts (\textit{e.g.}, reflection), \textit{etc.} During CNN model training, strong patterns make more contributions,  and the resultant model is more discriminative for them. However, if these patterns are absent in the testing data, the performance would severely drop. The CNN-based methods tend to overfit to some strong spoof patterns and thus suffer poor generalizability \cite{ij:liu2018learning}. Apart from overfitting, domain shift~\cite{ij:li2018unsupervised} is also an important reason for the poor generalizability of face anti-spoofing methods.  A domain  here refers to a certain environment where an image is acquisited, consisting of various factors such as illumination, background, facial appearance, camera type, \textit{etc.} Considering the huge diversity of real world environments, it is very common that  different samples have different domains. For example, the domains of two paper attacks may be quite different even in case of the same face if reproduced with different pieces of paper (\textit{e.g.} glossy \emph{vs.} rough paper). Such domain variance may lead to distribution dissimilarity of different samples in the feature space and cause the models to fail on new domains.

Based on the above observations, we propose a new Total Pairwise Confusion (TPC) loss to balance the contributions of all involved  spoof patterns, and also employ a Fast Domain Adaptation (FDA) model \cite{ij:engstrom2016faststyletransfer} to narrow the distribution discrepancy of samples from different domains in the feature space. We then obtain a Generalizable Face Authentication CNN model, shorted as GFA-CNN. Different from prior methods that take face anti-spoofing as a pre-step of face authentication, our GFA-CNN works in a multi-task manner, performing simultaneously face anti-spoofing and face recognition, as shown in Fig.~\ref{fig:0}.  Since the CNN layers of the two tasks share the same parameters, our model works with high efficiency.

Extensive experiments on five popular benchmarks for face anti-spoofing demonstrate the superiority of our method over the state-of-the-arts. Our code and trained models will be available upon acceptance. Our contributions are summarized as follows:
\begin{itemize}
\setlength{\itemsep}{0pt}
\setlength{\parsep}{0pt}
\setlength{\parskip}{0pt}
\item We propose a Total Pairwise Confusion (TPC) loss to effectively relieve the overfitting problems of CNN-based face anti-spoofing models to dataset-specific spoof patterns, which improves generalizability of face anti-spoofing methods.
\item We incorporate the Fast Domain Adaptation (FDA) model to learn more robust  Presentation Attack (PA) representations, which reduces  domain shift in the feature space.
\item We develop a multi-task CNN model for face authentication. Our GFA-CNN performs jointly face anti-spoofing and face recognition.
\end{itemize}

\section{Related Work}

Most previous approaches for face anti-spoofing exploit texture differences between live and spoofing faces with pre-defined features such as LBP \cite{ij:chingovska2012effectiveness}, HoG \cite{ij:komulainen2013context}, and SURF \cite{ij:boulkenafet2017face}, which are subsequently fed to a supervised classifier (\textit{e.g}., SVM, LDA) for binary classification. However, such handcrafted features are very sensitive to different illumination conditions, camera devices, specific identities, \textit{etc}. Though noticeable performance  achieved under the intra-dataset protocol, the sample from a different environment may fail the model. In order to obtain features with better generalizability, some approaches leverage temporal information, \textit{e.g.} making use of the spontaneous motions of the live faces, such as eye-blinking \cite{ij:pan2007eyeblink} and lip motion \cite{ij:kollreider2007real}. Though these methods are effective against photo attacks, they become vulnerable when attackers  simulate these motions through a paper with eye/mouth positions cut.

Recently, deep learning based methods \cite{ij:yang2014learn,ij:li2018learning} have been proposed to address face anti-spoofing. They use CNNs to learn highly discriminative representations by taking face anti-spoofing as a binary classification problem. However, most of them easily suffer overfitting. Current publicly available face anti-spoofing datasets are too limited to cover various potential spoofing types. A very recent work \cite{ij:liu2018learning} by Liu \textit{et al}. leverages the depth map and rPPG signal as auxiliary supervision to train CNN instead of treating face anti-spoofing as a simple binary classification problem in order to avoid overfitting. Another critical issue for face anti-spoofing is domain shift. To bridge the gap between training and testing domains, \cite{ij:li2018learning} generalizes CNN to unknown conditions by minimizing the feature distribution dissimilarity across domains, \textit{i.e}. minimizing the Maximum Mean Discrepancy distance among representations.

To our best knowledge, almost all previous works take face anti-spoofing as a pre-step prior to face recognition and address it as a binary classification problem. Compared with previous literature, we solve face anti-spoofing and face recognition at one shot. A most related work to ours is \cite{ij:sajjad2018cnn}, which proposed a two-tier framework to ensure the authenticity of the user to the recognition system, namely, monitoring whether the user has passed the biometric system as a live or spoofing one. It performs authentication based on fingerprint, palm vein print, face, \textit{etc.}, with two separated tiers: the anti-spoofing is powered by CNN learned representations while the recognition is based on pre-defined handcrafted features like ORB points.

Different with \cite{ij:sajjad2018cnn}, we build our GFA-CNN in a multi-task manner, our framework can recognize the identity of a given face, and meanwhile judge whether the face is a live or spoofing one. It is worth mentioning that for face recognition, our method achieves single-model accuracy up to 97.1\% on the LFW database \cite{ij:huang2008labeled}, which is even comparable to state-of-the-arts.

\begin{figure}
    \hspace{0.4cm}
    \includegraphics[width=7.3cm, height=3.2cm]{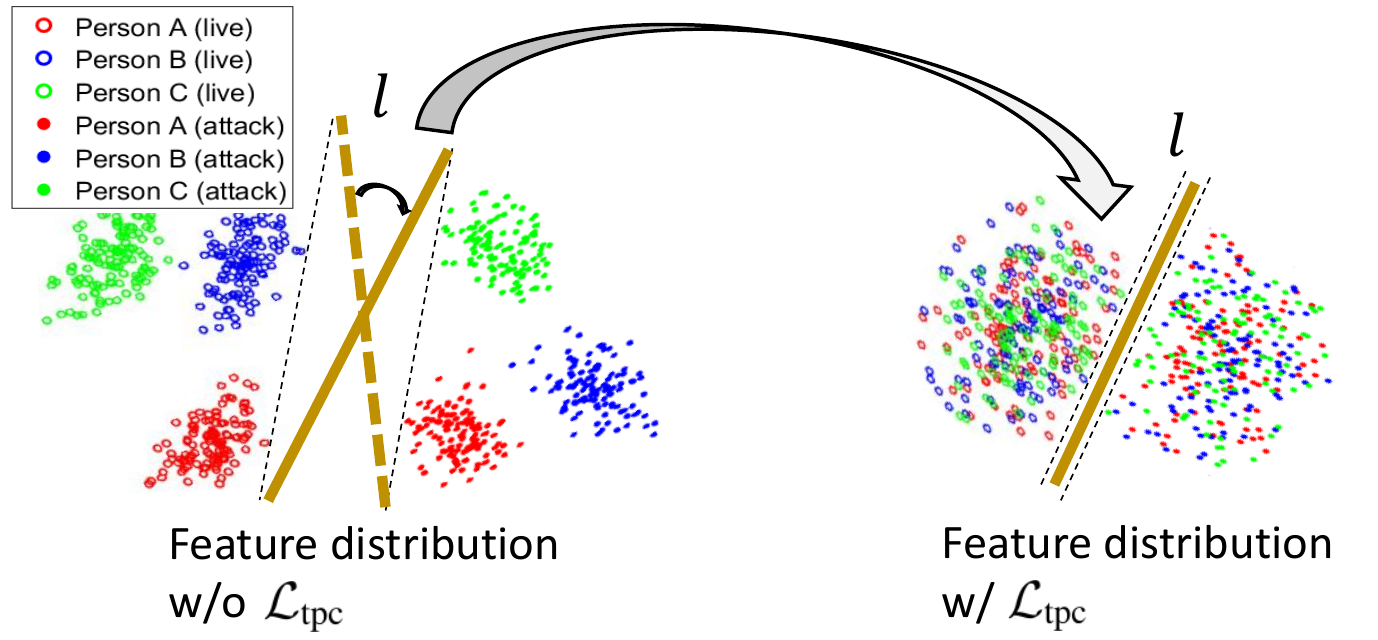}%
   \vspace{-0.2cm}
   \caption{Visualized comparison of the learned feature distribution w/ and w/o $\mathcal{L}_{\text{tpc}}$. Without $\mathcal{L}_{\text{tpc}}$, the feature distribution is diverse and person-specific (left), while with $\mathcal{L}_{\text{tpc}}$, the feature distribution becomes compact and homogeneous (right). $l$ is the classification hyperplane. Best viewed in color.}
   \vspace{-0.3cm}
\label{fig:2}
\end{figure}

\section{Generalizable Face Authentication CNN}
\subsection{Multi-Task Network Architecture}
The proposed Generalizable Face Authentication CNN (GFA-CNN) is able to jointly address face recognition and face anti-spoofing in a mutual boosting way. The network has two branches: the face anti-spoofing branch and the face recognition branch. Each branch consists of 5 blocks of CNN layers  and 3 fully connected (FC) layers, and each block contains 3 CNN layers. The parameters are shared between these two branches. The face anti-spoofing branch is trained by minimizing TPC loss and face anti-spoofing loss (Anti-loss), while the face recognition branch is trained by optimizing face recognition loss (Recg-loss). The anti-spoofing branch takes as input raw face images with background, while the recognition branch takes cropped faces as input. Before fed to the face anti-spoofing branch, the training images are transferred to a target domain by a given target-domain image. In testing phase, each query image is transferred to the target domain and then propagated forward the network.

\begin{figure}[t!]
    \hspace{0.3cm}
    \includegraphics[width=7.7cm, height=2.6cm]{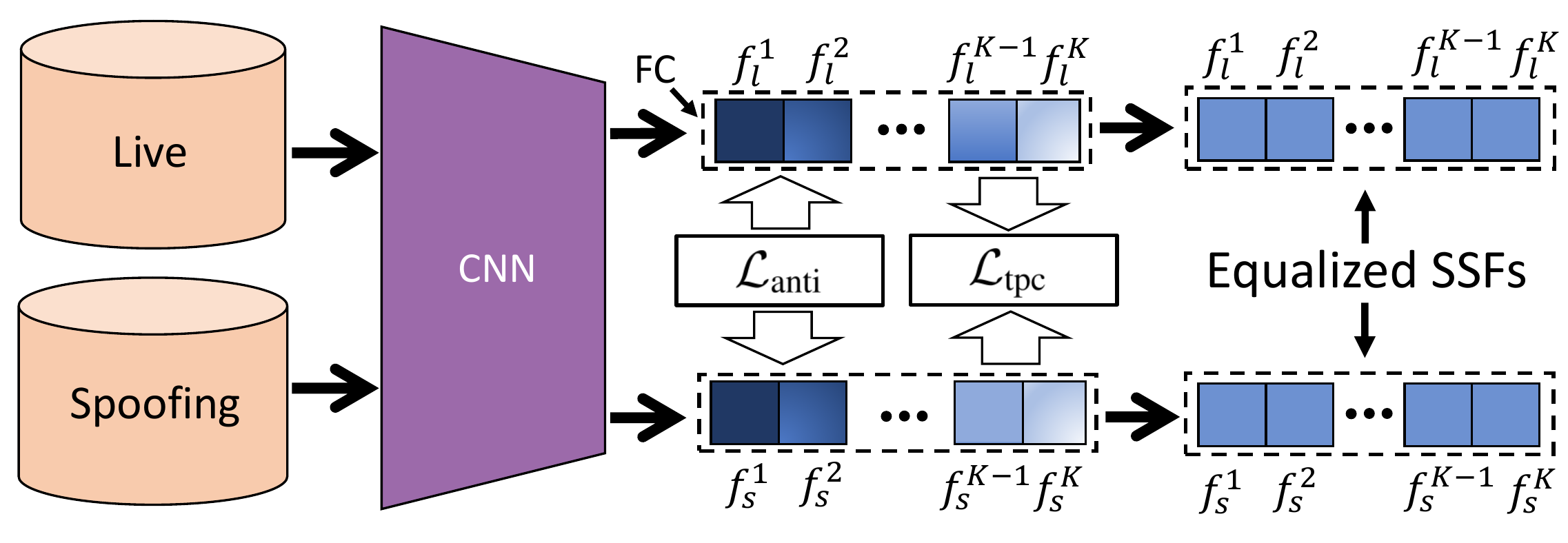}%
   \vspace{-0.2cm}
   \caption{The contribution-balanced process of SSFs. Darker color in the FC layer indicates a higher contribution to the classification while lighter color indicates lower. Each grid represents an SSF. The trade-off game between $\mathcal{L}_{\text{tpc}}$ and $\mathcal{L}_{\text{anti}}$ can balance the contributions of SSFs to the final decision.}
   \vspace{-0.3cm}
\label{fig:3}
\end{figure}

The CNN blocks are structured the same with the convolution part of VGG16. Before training, the CNN blocks are first trained on the VGG-face dataset to obtain fundamental weights for face recognition. The FC layers of face anti-spoofing and face recognition branches have the same structure except for the output dimension of the last FC layer. The face anti-spoofing branch takes 2 dimensions for the last FC layer, while the dimensions of the last FC layer in the face recognition branch depend on the number of subjects involved in training. The overall objective function is
\begin{equation}
\small
\begin{aligned}
\mathcal{L} = \mathcal{L}_{\text{anti}} + \lambda_1*\mathcal{L}_{\text{id}} + \lambda_2*\mathcal{L}_{\text{tpc}},
\end{aligned}
\end{equation}
where $\mathcal{L}_{\text{anti}}$ and $\mathcal{L}_{\text{recg}}$ are the cross entropy losses for face anti-spoofing and face recognition respectively, $\mathcal{L}_{\text{tpc}}$ is the Total Pairwise Confusion (TPC) loss, and $\lambda_1$ and $\lambda_2$ are the weighting parameters among different losses.

\subsection{Total Pairwise Confusion Loss}
In order to learn Presentation Attack (PA) representations that are adaptable to varying environment conditions, we propose a novel Total Pairwise Confusion (TPC) loss. Our inspiration comes from the pairwise confusion (PC) loss~\cite{ij:dubey2018pairwise} that  tackles the overfitting issue in fine-grained visual classification by intentionally introducing confusion in the feature activations.  We modify their confusion implementation to make it applicable to the face anti-spoofing task. Our TPC loss is defined as
\begin{equation}
\small
\begin{aligned}
\mathcal{L}_{tpc}(\textbf{x}_i, \textbf{x}_j) = \sum_{i \neq j}^{M}||\psi(\textbf{x}_i) - \psi(\textbf{x}_j)||_2^2,
\end{aligned}
\end{equation}
where $\textbf{x}_i$ and $\textbf{x}_j$ are two randomly selected images (sample pair), $M$ is the total number of sample pairs involved in training and $\psi(\textbf{x})$ denotes the representations of the second fully connected layer of the face anti-spoofing branch.

Our $\mathcal{L}_{\text{tpc}}$ differs from the original PC loss in two-fold: 1) TPC loss minimizes the distribution distance of a random sample pair from the training set, rather than the sample pair from two different categories, to force CNN to learn slightly less discriminative features. 2) We minimize the Euclidean distance in the feature space while the original PC loss minimizes the distance in the probability space (output of softmax) to make samples in the same pair have a similar conditional probability distribution.

Our modifications are based on below considerations: 1) With face anti-spoofing taken as a binary classification issue, confusion across categories would not excessively affect the discriminability of the PA feature on differentiating live \textit{vs}. spoofing samples. 2) Face samples related to the same subject  would usually cluster in the feature space, and implementing confusion on all samples could compact and homogenize the whole feature distribution (see Fig.~\ref{fig:2}), thus benefiting generalization performance. 3) As a binary classification problem of simpler structure, regularizing the model within the feature space would be more useful than imposing regularization within the output probabilistic space.

Our $\mathcal{L}_{\text{tpc}}$ can effectively improve the generalizability of PA representations. This can be understood as follows. Suppose there are $K$ components in the PA representations, each corresponding to one spoof pattern, which is called a Spoof-pattern Specific Feature (SSF) in this work. As shown in Fig.~\ref{fig:3},  different SSFs contribute differently to the final decision. If we define the feature for a live and a spoofing sample as $F_l = (\textbf{f}_l^1, \textbf{f}_l^2, ..., \textbf{f}_l^K)$ and $F_s = (\textbf{f}_s^1, \textbf{f}_s^2, ..., \textbf{f}_s^K)$, respectively, where $\textbf{f}_l^i$ is the $i^{th}$ SSF of the live sample and $\textbf{f}_s^{i}$ is the $i^{th}$ SSF of the spoofing sample. The SSFs are ranked based on their importance to the classification of live \textit{vs.} spoofing. On one hand, $\mathcal{L}_{\text{anti}}$ aims to enlarge the distance between $F_l$ and $F_s$ for better discrimination. On the other hand, $\mathcal{L}_{\text{tpc}}$ attempts to narrow the difference between $F_l$ and $F_s$. As $\textbf{f}^1_{l/s}$ contributes the most to the differentiation of live and spoofing samples, it will be impaired the most by $\mathcal{L}_{\text{tpc}}$. However, the contributions of less important SSFs, such as $\textbf{f}_{l/s}^{K-1}$ and $\textbf{f}_{l/s}^K$, will be enhanced by $\mathcal{L}_{\text{anti}}$ to offset the impaired discriminative ability. In this trade-off game, the contributions of all SSFs tend to be equalized, meaning more spoof patterns are involved in the decision rather than just a couple of strong spoof patterns specific to the training set. This could effectively alleviate overfitting risks. If some spoof patterns disappear in testing, a fair decision can still be achieved by other patterns, ensuring CNN would not  overfit to some specific features.

\subsection{Fast Domain Adaptation}
Besides the proposed TPC loss that balances the contribution of each spoof pattern, we also apply FDA to reduce domain shift in the feature space to further improve the generalizability of our framework.

Generally, an image contains two components: content and appearance \cite{ij:pan2018two}. The appearance information (\textit{e.g.}, colors, localised structures) makes up the style of images from a certain domain and is mostly represented by features in the bottom layers of CNN \cite{johnson2016perceptual}. For face anti-spoofing, the domain variance among face samples may introduce the distribution dissimilarity in the feature space and hurt anti-spoofing performance. Here, we employ the FDA to alleviate negative effects brought by domain changes. The FDA consists of  an image transformation network $f(\cdot)$ that generates a synthetic image $y$ from a given image $x$: $y = f(x)$, and a loss network $\varphi(\cdot)$ that computes content reconstruction loss $\mathcal{L}_{\text{content}}$ and domain reconstruction loss $\mathcal{L}_{\text{domain}}$.

Let $\varphi_j(\cdot)$ be the $j^{th}$ layer of $\varphi(\cdot)$ with the shape of $C_j \times H_j \times W_j$. The content reconstruction loss penalizes the output image $y$ when it deviates in content from the input $x$. We thus minimize the Euclidean distance between the feature representations of $y$ and $x$:
\begin{equation}
\small
\begin{aligned}
\mathcal{L}_{\text{content}} = \frac{1}{C_jH_jW_j}||\varphi_j(y) - \varphi_j(x)||_2^2.
\end{aligned}
\end{equation}
The domain reconstruction loss enables the output image $y$ to have the same domain with the target-domain image $y_d$. We then minimize the squared Frobenius norm of the difference between the Gram matrices of $y$ and $y_d$:
\begin{equation}
\small
\begin{aligned}
\mathcal{L}_{\text{domain}} = \frac{1}{C_jH_jW_j}||G_j(y) - G_j(y_d)||_F^2.
\end{aligned}
\end{equation}
The Gram matrix is computed by reshaping $\varphi_j$ into a matrix $\kappa$, $G_j = \kappa\kappa^{T}/C_jH_jW_j$. Then the optimal image $\hat{y}$ is generated by solving the following objective function:
\begin{equation}
\small
\begin{aligned}
\hat{y} = \mathop{\arg\min}_{P}(\lambda_c \mathcal{L}_{\text{content}}(y, x) + \lambda_s\mathcal{L}_{\text{domain}}(y, y_d)),
\end{aligned}
\end{equation}
where $P$ is the optimal parameters of network $f(\cdot)$, $x$ is the content image, $y=f(x)$, $y_d$ is the target-domain image, and $\lambda_c$ and $\lambda_s$ are scalars. By solving Eqn. (5), $x$ is transferred to $\hat{y}$, preserving the content of $x$ with the domain of $y_d$.

\begin{figure}
    \hspace{0.1cm}
    \includegraphics[width=8.0cm, height=4.3cm]{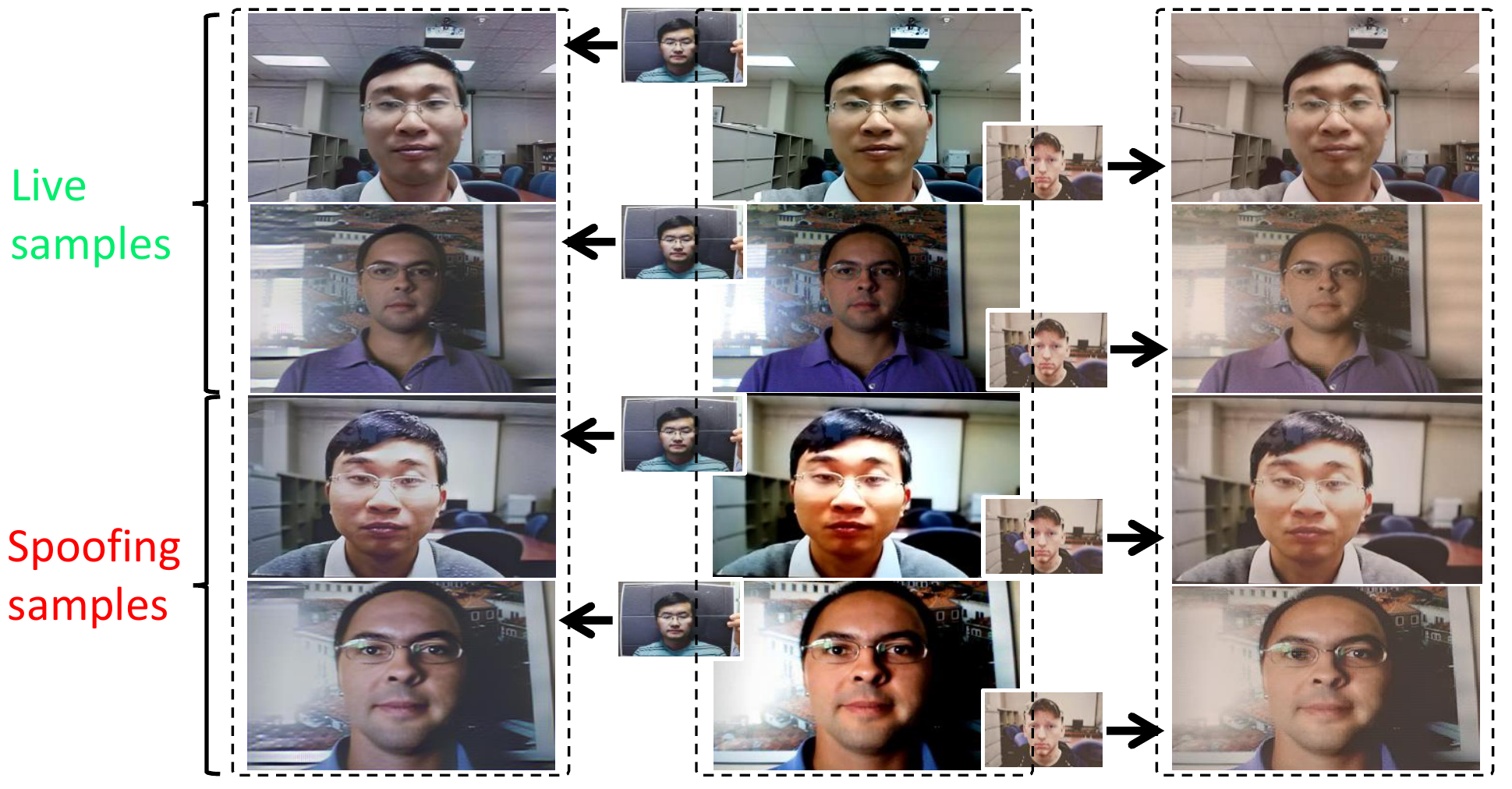}%
   \vspace{-0.2cm}
   \caption{Example results by FDA. The upper left and bottom right images of the images in the middle column are the target-domain images expected to be transferred. Images of odd rows are from MSU-MFSD; images of even rows are from Replay-Attack.}
   \vspace{-0.3cm}
\label{fig:4}
\end{figure}

Fig.~\ref{fig:4} shows some of our domain transferred samples. The target-domain image is sampled from the training data. Detailed analysis on the feature diversity between domains w/ and w/o FDA is provided in Sec. 4.2.

\section{Experiments}
\subsection{Experimental Setup}
\paragraph{Datasets.} We evaluate GFA-CNN  on five face anti-spoofing benchmarks: CASIA-FASD \cite{ij:zhang2012face}, Replay-Attack \cite{ij:chingovska2012effectiveness}, MSU-MFSD \cite{ij:wen2015face}, Oulu-NPU \cite{ij:boulkenafet2017oulu} and SiW \cite{ij:liu2018learning}. CASIA-FASD and MSU-MFSD are small datasets, containing 50 and 35 subjects, respectively. Oulu-NPU and SiW are high-resolution databases published very recently. Oulu-NPU contains 4 testing protocols: Protocol 1 evaluates the environment condition variations; Protocol 2 examines the influences of different spoofing mediums; Protocol 3 estimates the effects of different input cameras; Protocol 4 considers all the challenges above. We conduct intra-database tests on MSU-MFSD and Oulu-NPU, respectively. Cross-database tests are performed between CASIA-FASD \textit{vs.} Replay-Attack and MSU-MFSD \textit{vs.} Replay-Attack, respectively. The face recognition performance is evaluated on SiW, which contains 165 subjects with large variations in poses, illumination, expressions (PIE), and different distances from subject to camera. The  LFW,  the most widely used benchmark for face recognition, is also used to evaluate the face recognition performance.

\paragraph{Implementation Details.} The proposed GFA-CNN is implemented with TensorFlow \cite{ij:abadi2016tensorflow}. We use Adam optimizer with a learning rate beginning at 0.0003 and decaying half after every 2,000 steps. The batch size is set as 32.  $\lambda_1$ and $\lambda_2$ in Eqn. (1) are set as 0.1 and $2.5e^{-5}$, respectively.  All experiments are performed according to the protocols provided in the datasets. The CNN layers are pre-trained on the VGG-face dataset \cite{ij:parkhi2015deep}. For data balance, we triple the live samples in the training set of CASIA-FASD, MSU-MFSD and Replay-Attack with horizontal and vertical flipping, while doubling the live samples in the training set of SiW by just flipping horizontally.

\paragraph{Evaluation Metrics.} We have two evaluation protocols, intra-test and cross-test, which test samples from and not from the domain of the training set, respectively. We report our results with the following metrics. Intra-test evaluation: Equal Error Rate (EER), Attack Presentation Classification Error Rate (APCER), Bona Fide Presentation Classification Error Rate (BPCER) and, ACER=(APCER+BPCER)/2. Cross-test evaluation: HTER.

\begin{table}
\small
\begin{center}
\resizebox{0.48 \textwidth}{!}{
\begin{tabular}{|c|c|c|c|c|}
\hline
\multirow{2}*{TPC/FDA} & \multicolumn{2}{c|}{Intra-Test} & \multicolumn{2}{c|}{Cross-Test} \\
\cline{2-5}
        & MFSD & Replay & MFSD $\rightarrow$ Replay& Replay $\rightarrow$ MFSD \\
\hline\hline
$-$\quad$-$ & 10.5 & 0.6 & 39.4 & 34.6\\
$-$\quad$+$ & 11.2 & 0.6 & 36.3 & 38.3\\
$+$\quad$-$ & \textbf{6.4} & \textbf{0} & 28.5 & 26.6\\
$+$\quad$+$ & 8.3 & 0.3 & \textbf{25.8} & \textbf{23.5}\\
\hline
\end{tabular}}
\end{center}
\vspace{-0.2cm}
\caption{Ablation study (HTER \%). бобо+'' means the corresponding component is used, while бобо-'' indicates removing the component. The numbers in bold are the best results.}
\end{table}

\subsection{Ablation Study}
We perform ablation analysis to reveal the role of TPC loss and FDA in our framework. We retrain the proposed network by adding/ablating TPC and FDA. As shown in Tab. 1, if TPC is removed, the HTER of intra-test on MFSD drops by 2.9\% (w/ FDA) and 4.1\% (w/o FDA), respectively. Since Replay-Attack is usually free of severe overfitting, it is reasonable to see the improved performance is not significant when using FDA, 0.3\% (w/ FDA) and 0.6\% (w/o FDA) on HTER.

For cross-test, if TPC is ablated, the HTER  dramatically decreases by over 10\% for MFSD $\rightarrow$ Replay\footnote{The acronym $*$ $\rightarrow$ $\diamond$ means training on database бобо$*$'' and testing on database бобо$\diamond$''.}, and over 8\% for Replay $\rightarrow$ MFSD, no matter FDA is used or not. The best cross-test result is achieved by using both TPC and FDA, indicating FDA can further improve the generalizability of the proposed method.

\begin{figure}
    \hspace{0.0cm}
    \includegraphics[width=8.5cm, height=3.1cm]{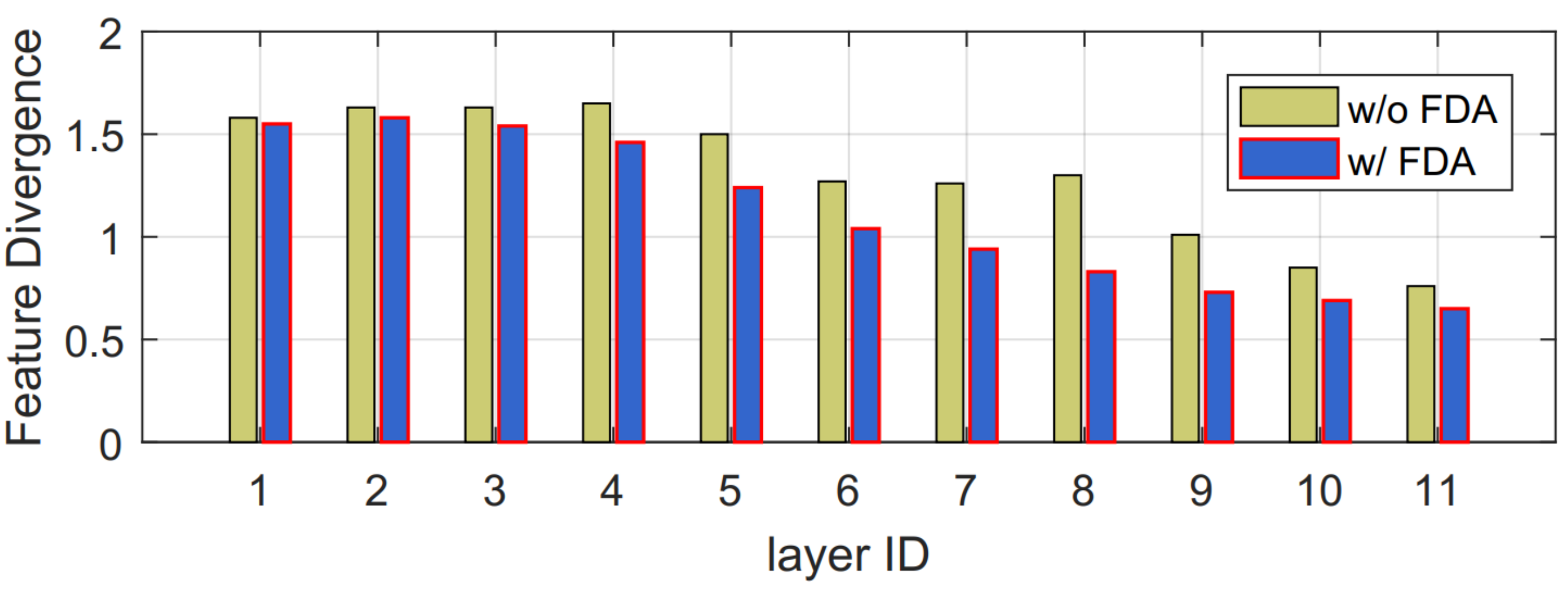}%
   \vspace{-0.2cm}
   \caption{Feature divergence comparison between MSU-MFSD and Replay-Attack. The numbers on x-axis correspond to the CNN layer of VGG16.}
   \vspace{-0.15cm}
\label{fig:7}
\end{figure}

To evaluate the feature diversity between domains w/ and w/o FDA, we calculate the feature divergence via symmetric KL divergence. Similar to \cite{ij:pan2018two}, we denote the mean value of a channel from the feature embedding of CNN as $F$. Given a Gaussian distribution of $F$, with mean $\mu$ and variance $\sigma^2$, the symmetric KL divergence of this channel between domain A and B is
\begin{equation}
\small
\begin{aligned}
D(F_A||F_B) = KL(F_A||F_B) + KL(F_B||F_A).
\end{aligned}
\end{equation}
\begin{equation}
\small
\begin{aligned}
KL(F_A||F_B) = log\frac{\sigma_A}{\sigma_B} + \frac{\sigma_A^2 + (\mu_A - \mu_B)^2}{2\mu_B^2} - \frac{1}{2}.
\end{aligned}
\end{equation}
Denote $D(F_{iA}||F_{iB})$ as the symmetric KL divergence of the $i^{th}$ channel. Then the average feature divergence of the layer is defined as
\begin{equation}
\small
\begin{aligned}
D(L_A||L_B) = \frac{1}{C}\sum_{i=1}^{C}D(F_{iA}||F_{iB}),
\end{aligned}
\end{equation}
where C is the channel number of this layer. This metric measures the distance between the feature distributions of domain A and B. We calculate the feature divergence of each layer in a CNN model for comparison. In particular, we randomly select 5,000 face samples from MSU-MFSD and Replay-Attack, respectively. Each dataset is considered as one domain. These samples are then fed to a pre-trained VGG16 \cite{ij:simonyan2014very} model to calculate the KL divergence at each layer following Eqn. (8). The comparison results are shown in Fig.~\ref{fig:7}. As can be seen, with the FDA, the feature divergence between MSU-MFSD and Replay-Attack is significantly reduced.
\begin{table}
\small
\begin{center}
\resizebox{0.35\textwidth}{!}{
\begin{tabular}{|l|c|c|c|}
\hline
 Methods & EER(\%)  \\
\hline\hline
 LBP + SVM baseline & 14.7 \\
 DoG + LBP + SVM baseline & 23.1 \\
 IDA + SVM \cite{ij:wen2015face} & 8.58 \\
 Color LBP \cite{ij:boulkenafet2015face} & 10.8 \\
 Color texture \cite{ij:boulkenafet2016face} & 4.9 \\
 Color SURF \cite{ij:boulkenafet2017face} & \textbf{2.2} \\
 GFA-CNN (ours)   & 7.5 \\
\hline
\end{tabular}}
\end{center}
\vspace{-0.2cm}
\caption{Intra-test results on MSU-MFSD. The numbers in bold are the best results.}
\end{table}

\begin{table}
\small
\begin{center}
\resizebox{0.46\textwidth}{!}{
\begin{tabular}{|c|l|c|c|c|}
\hline
Prot. & Methods & APCER(\%) & BPCER (\%) & ACER (\%) \\
\hline\hline
 & CPqD & 2.9 & 10.8 & 6.9\\
1 & GRADIANT & \textbf{1.3} & 12.5 & 6.9\\
 & GFA-CNN (ours) & 2.5 & \textbf{8.9} & \textbf{5.7}\\
\hline\hline
 & MixedFASNet & 9.7 & 2.5 & 6.1\\
2 & GRADIANT & 3.1 & 1.9 & 2.5\\
 & GFA-CNN (ours) & \textbf{2.5} & \textbf{1.3} & \textbf{1.9}\\
\hline\hline
 & MixedFASNet & 5.3 & 7.8 & 6.5\\
3 & GRADIANT & \textbf{2.6} & \textbf{5.0} & \textbf{3.8}\\
 & GFA-CNN (ours) & 4.3 & 7.1 & 5.7\\
\hline\hline
 & Massy HNU & 35.8 & 8.3 & 22.1\\
4 & GRADIANT & \textbf{5.0} & 15.0 & 10.0\\
 & GFA-CNN (ours) & 7.4 & \textbf{10.4} & \textbf{8.9}\\
\hline
\end{tabular}}
\end{center}
\vspace{-0.2cm}
\caption{Intra-test results on the four protocols of Oulu-NPU. The numbers in bold are the best results.}
\vspace{-0.1cm}
\end{table}

\begin{table*}
\small
\begin{center}
\resizebox{0.95\textwidth}{!}{
\begin{tabular}{|l|c|c|c|c|c|c|c|c|c|}
\hline
\multirow{2}*{Methods}&Train&Test&Train&Test&Train&Test&Train&Test&\multirow{2}*{Average} \\
\cline{2-9}
 & CASIA & Replay & Replay & CASIA & MFSD & Replay & Replay & MFSD & \\
 \hline
 LBP \cite{ij:de2013can} &\multicolumn{2}{c|}{47.0}&\multicolumn{2}{c|}{39.6}& \multicolumn{2}{c|}{45.5} & \multicolumn{2}{c|}{45.8} & 44.5\\
 \hline
 LBP-TOP \cite{ij:de2013can} &\multicolumn{2}{c|}{49.7}&\multicolumn{2}{c|}{60.6}& \multicolumn{2}{c|}{46.5} & \multicolumn{2}{c|}{47.5} & 51.1\\
\hline
 Motion \cite{ij:de2013can} &\multicolumn{2}{c|}{50.2}&\multicolumn{2}{c|}{47.9}& \multicolumn{2}{c|}{-} & \multicolumn{2}{c|}{-} & 49.1\\
\hline
 CNN \cite{ij:yang2014learn} &\multicolumn{2}{c|}{48.5}&\multicolumn{2}{c|}{45.5}& \multicolumn{2}{c|}{37.1} & \multicolumn{2}{c|}{48.6} & 44.9\\
\hline
  Color LBP \cite{ij:boulkenafet2018generalization} & \multicolumn{2}{c|}{37.9} & \multicolumn{2}{c|}{35.4} & \multicolumn{2}{c|}{44.8} & \multicolumn{2}{c|}{33.0} & 37.8\\
\hline
 Color Tex. \cite{ij:boulkenafet2018generalization} &\multicolumn{2}{c|}{30.3}&\multicolumn{2}{c|}{37.7}& \multicolumn{2}{c|}{33.9} & \multicolumn{2}{c|}{34.1} & 34.0\\
\hline
  Color SURF \cite{ij:boulkenafet2018generalization} & \multicolumn{2}{c|}{26.9} & \multicolumn{2}{c|}{\textbf{23.2}} & \multicolumn{2}{c|}{29.7} & \multicolumn{2}{c|}{31.8} & 27.9\\
\hline
 Auxiliary \cite{ij:jourabloo2018face} &\multicolumn{2}{c|}{27.6}&\multicolumn{2}{c|}{28.4}& \multicolumn{2}{c|}{-} & \multicolumn{2}{c|}{-} & 28.0\\
\hline
 De-Spoof \cite{ij:liu2018learning}&\multicolumn{2}{c|}{28.5}&\multicolumn{2}{c|}{41.1} & \multicolumn{2}{c|}{-} & \multicolumn{2}{c|}{-} & 34.8\\
\hline
GFA-CNN (ours) & \multicolumn{2}{c|}{\textbf{21.4}} & \multicolumn{2}{c|}{34.3} & \multicolumn{2}{c|}{\textbf{25.8}} & \multicolumn{2}{c|}{\textbf{23.5}} & \textbf{26.3}\\
\hline
\end{tabular}}
\end{center}
\vspace{-0.2cm}
\caption{Cross-test results (HTER $\%$) on CASIA-FASD, Replay-Attack, and MSU-MFSD. бобо-'' indicates the corresponding result is unavailable. The numbers in bold are the best results.}
\end{table*}

\subsection{Face Anti-spoofing Evaluation}
\paragraph{Intra-Test.} We perform intra-test on MSU-MFSD and Oulu-NPU. Tab. 2 shows the comparisons of our method with other state-of-the-art methods on MSU-MFSD. For Oulu-NPU, we refer to the face anti-spoofing competition results in \cite{ij:boulkenafet2017competition} and use the best two for each protocol for comparison. All results are reported in Tab. 3.

As shown in Tab. 2, GFA-CNN achieves the EER of 7.5\%, ranking the $3^{rd}$ among all the compared methods. This result is satisfactory considering GFA-CNN is not designed blindly to pursue high performance in the intra-test setting. In our experiments, we find the proposed TPC loss may slightly decrease the intra-test performance, mainly because TPC loss impairs the contributions of several strongest SSFs w.r.t the training datasets. The weakening of these dataset-specific features may in turn affect the intra-test performance (however, they may improve the performance in cross-test). According to Tab. 3, our method achieves the lowest ACER in 3 out of 4 protocols. For the most challenging protocol 4, we achieve the ACER of 8.9\%, which is 1.1\% lower than the best performer.

\begin{figure}
    \includegraphics[width=8.2cm, height=2.75cm]{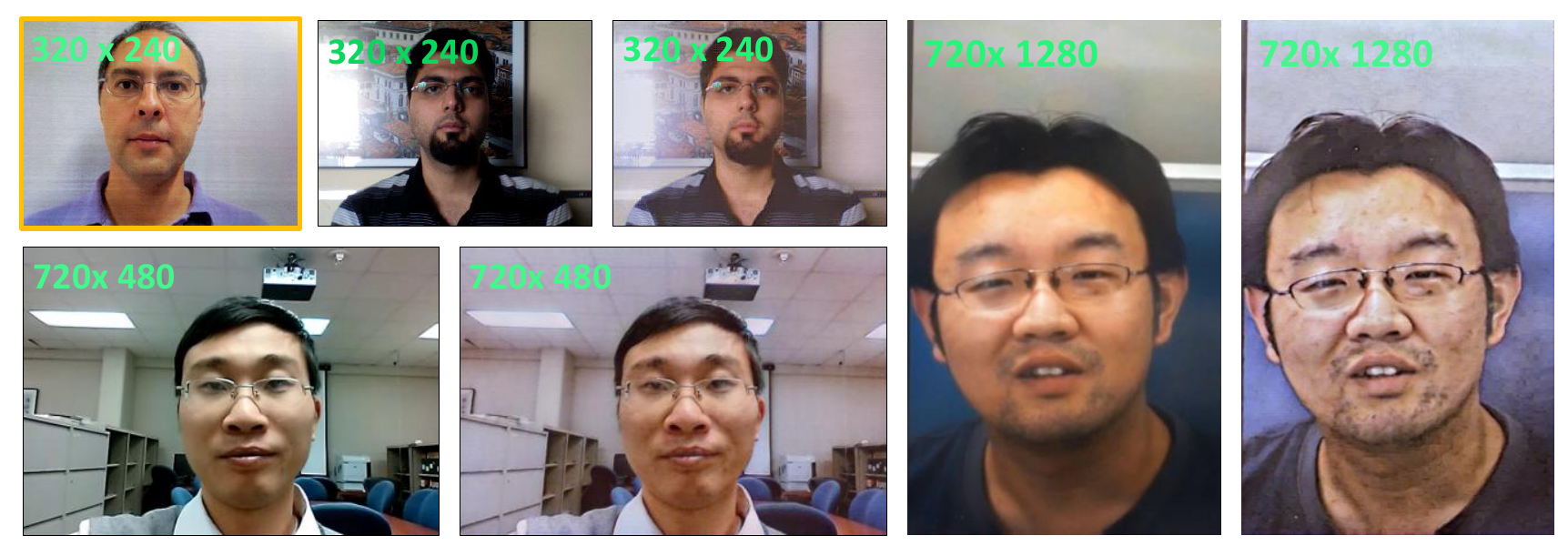}%
   \vspace{-0.2cm}
   \caption{Results transferred by FDA with different resolutions. The top left image is the target-domain image. For other images of each block, the left one is the original image, and the right is the transferred image. The green number located at the top left of each image indicates the resolution.}
   \vspace{-0.3cm}
\label{fig:5}
\end{figure}

\paragraph{Cross-Test.} To demonstrate the strong generalizability of GFA-CNN, we perform cross-test on CASIA-FASD, Replay-Attack, and MSU-MFSD by comparing with other state-of-the-arts. We adopt the most widely used cross-test settings: CASIA-FASD \textit{vs.} Replay-Attack and MSU-MFSD \textit{vs.} Replay-Attack, and report  comparison results  in Tab. 4. As can be seen, GFA-CNN achieves the lowest HTERs in cross-test: CASIA $\rightarrow$ Replay, MFSD $\rightarrow$ Replay and Replay $\rightarrow$ MFSD. Especially for Replay $\rightarrow$ MFSD, GFA-CNN reduces the cross-testing HTER by 8.3\% compared with the best state-of-the-art.

However, we also observe  GFA-CNN has a relatively worse HTER compared with the best method on Replay Attack $\rightarrow$ CASIA-FASD. This  is probably due to the ``quality degradation" by FDA when the resolution of a source-domain image to be transferred is much higher than that of the target-domain image. During the cross-testing on Replay-Attack $\rightarrow$ CASIA-FASD, the target-domain image is selected from Replay-Attack with a low-resolution of $320\times240$. However, CASIA-FASD contains quite a number of images with high-resolution of $720\times1280$. Such a ``resolution gap" leads to a ``quality degradation" of FDA, as shown in the rightmost image in Fig.~\ref{fig:5}.

\subsection{Face Recognition Evaluation}
We further evaluate the face recognition performance of our GFA-CNN on  SiW and LFW. Since our method is not targeted specifically at face recognition, we only adopt VGG-16 as the baseline. On LFW, we follow the provided protocol to perform testing.
On SiW we use  90 subjects for training and the other 75 subjects for testing, which is its default data splitting. This dataset also provides a frontal legacy face image corresponding to each subject. At the testing phase, we select the legacy image w.r.t each subject of the testing set as the gallery faces, and use all images in the testing set (including both live and spoofing) as the probe faces.
\begin{figure}
    \hspace{0.0cm}
    \includegraphics[width=8.2cm, height=3.0cm]{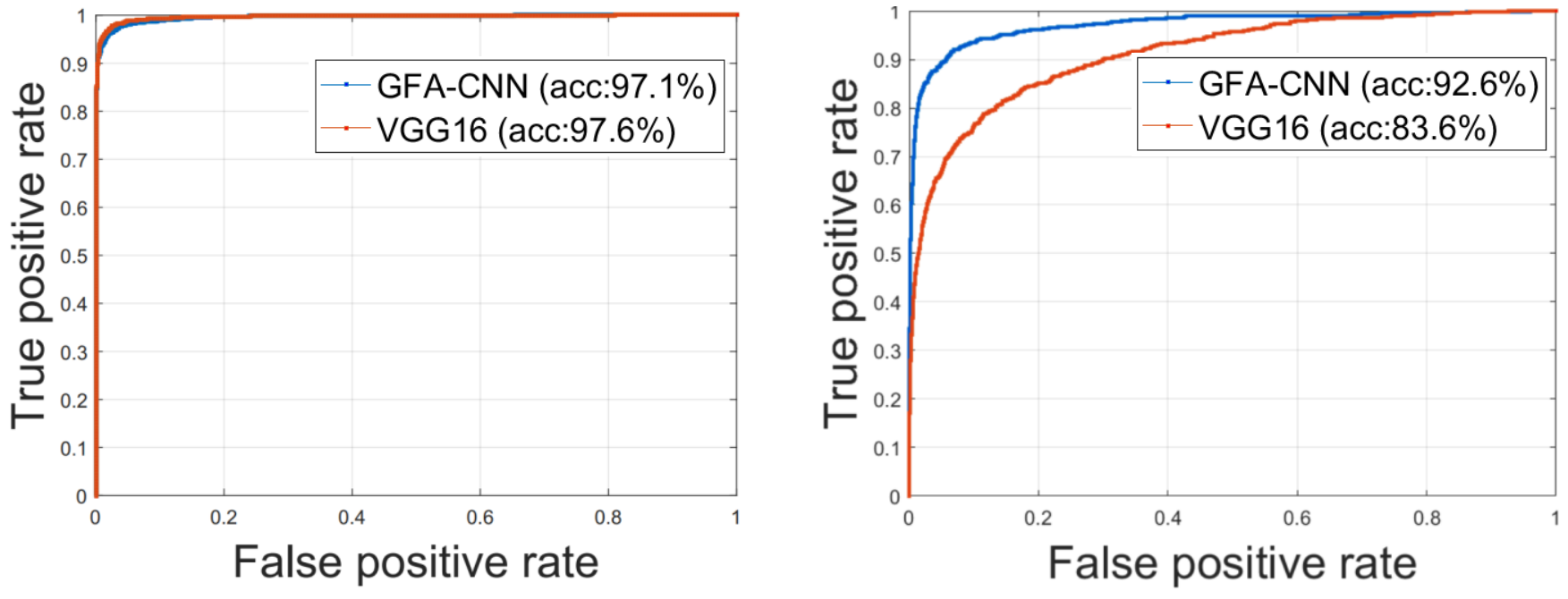}%
   \vspace{-0.2cm}
   \caption{Comparison of ROC curves of face verification on (a) LFW and (b) SiW.}
   \vspace{-0.3cm}
\label{fig:6}
\end{figure}

The ROC curves of face verification are shown in Fig.~\ref{fig:6}. As can be observed, GFA-CNN achieves competitive results to VGG16 on LFW, 97.1\% and 97.6\%, respectively. However, when testing on SiW, the declined accuracy of GFA-CNN is much lower than that of VGG16: the accuracy of GFA-CNN reduces by 4.5\%, while VGG16 drops by 14\%. The degraded performance is mainly due to face reproduction by spoofing mediums, in which some of the finer facial details might be lost. However, GFA-CNN still achieves satisfactory performance compared with VGG16. This is mainly because the face anti-spoofing and face recognition tasks mutually enhance each other, making the representations learned for face recognition less sensitive to spoof patterns.

\subsection{Discussions on Multi-task Setting}
In this subsection, we investigate how the multi-task learning affects model performance for face anti-spoofing. We retrain our model without the face recognition branch, keep hyper-parameters unchanged and evaluate with the same protocol as the GFA-CNN. From the experiments, we observe the multi-task training slightly decreases the intra-test performance of face anti-spoofing (dropping 2.5\% and 0.3\% on MSU-MFSD and Replay-Attack, respectively). This is reasonable, since the single model learns to perform two different tasks. However, two advantages are achieved compared with the single task training. Firstly, the training process becomes more stable with the Anti-loss decreasing gradually, rather than dropping sharply after some steps by single task training, suggesting multi-task setting can help overcome overfitting. Secondly, as shown in Fig.~\ref{fig:6}, multi-task training helps learn face representations less sensitive to spoof patterns for face recognition. This mainly benefits from sharing parameters in the convolutional layers, giving more generic fusion features.

\section{Conclusion}
This paper presents a novel CNN model to jointly address face recognition and face anti-spoofing in a mutual boosting way.
In order to learn more generalizable Presentation Attack (PA) representations for face anti-spoofing, we propose a novel Total Pairwise Confusion (TPC) loss to balance the contribution of each spoof pattern, preventing the PA representations
from overfitting to dataset-specific spoof patterns. The Fast Domain Adaptation (FDA) is also incorporated into our framework to reduce distribution dissimilarity of face samples from different domains, further enhancing the robustness of PA representations. Extensive experiments on both face anti-spoofing and face recognition datasets show that our GFA-CNN achieves not only superior performance for face anti-spoofing on cross-tests, but also high accuracy for face recognition.

\section*{Acknowledgement}

{\small
\bibliographystyle{ieee}
\bibliography{ijcai19_byZJ}
}

\end{document}